\documentclass{article}
\usepackage{ijcai22}

\usepackage{times}
\usepackage{soul}
\usepackage{url}
\usepackage[hidelinks]{hyperref}
\usepackage[utf8]{inputenc}
\usepackage[small]{caption}
\usepackage{graphicx}
\usepackage{amsmath}
\usepackage{amsthm}
\usepackage{booktabs}
\usepackage{algorithm}
\usepackage{algorithmic}
\usepackage{amsfonts}
\usepackage{xcolor}

\usepackage{lipsum}
\usepackage[margin=1in]{geometry}
\usepackage{multicol, multirow, makecell}
\usepackage{booktabs}

\title{Off-Policy Reinforcement Learning with Loss Function Weighted by Temporal Difference Error}

\author{
Bumgeun Park$^1$
\and
Taeyoung Kim$^1$\and
Woohyeon Moon$^1$\and
Luiz Felipe Vecchietti$^2$\And
Dongsoo Har$^1$
\affiliations
$^1$Cho Chun Shik Graduate School of Mobility, KAIST\\\
$^2$Data Science Group, Institute for Basic Science\\
\emails
\{j4t123, ngng9957, moonstar, dshar\}@kaist.ac.kr,
lfelipesv@ibs.re.kr
}

\begin{document}

\maketitle

\begin{abstract}
Training agents via off-policy deep reinforcement learning (RL) requires a large memory, named replay memory, that stores past experiences used for learning. These experiences are sampled, uniformly or non-uniformly, to create the batches used for training. When calculating the loss function, off-policy algorithms assume that all samples are of the same importance. In this paper, we hypothesize that training can be enhanced by assigning different importance for each experience based on their temporal-difference (TD) error directly in the training objective. We propose a novel method that introduces a weighting factor for each experience when calculating the loss function at the learning stage. In addition to improving convergence speed when used with uniform sampling, the method can be combined with prioritization methods for non-uniform sampling. Combining the proposed method with prioritization methods improves sampling efficiency while increasing the performance of TD-based off-policy RL algorithms. The effectiveness of the proposed method is demonstrated by experiments in six environments of the OpenAI Gym suite. The experimental results demonstrate that the proposed method achieves a 33\%$\sim$76\% reduction of convergence speed in three environments and an 11\% increase in returns and a 3\%$\sim$10\% increase in success rate for other three environments.
\end{abstract}

\section{Introduction}
Reinforcement learning (RL) enables an agent to learn a task by taking action in an environment so as to maximize future rewards \cite{RL}.
The use of deep neural networks for the nonlinear approximation of a policy or value function allows RL algorithms to deal with complicated tasks, such as playing a range of Atari games and mastering the game of Go without human knowledge \cite{atrai,go1,go2,deep_rl}.
Recently, there has also been some research on the application of deep RL not only in game environments but also in practical environments. Examples include controlling a robotic arm \cite{lab1,lab2,lab3}, planning a path for mobile robots \cite{lab4,lab5}, controlling soccer robots in a cooperative-competitive environment \cite{lab6,lab7,lab8}, and predicting traffic accidents \cite{predicting_accident}. 
One of these challenges is strongly correlated updates that break the i.i.d. assumption, which is central to almost all aspects of machine learning.


Experience replay (ER) is one of the elements of off-policy deep RL \cite{ER}, and it can solve strongly correlated updates by allowing all experiences to be uniformly sampled.
Also, because ER uses past experience, the sample efficiency is increased.
The agent stores experience in memory, known as replay memory, at every timestep. 
Then, ER randomly recalls experiences as a mini-batch from memory to train a neural network, decreasing the correlation between experiences. 
However, there are major challenges associated with ER, such as the paucity of successful experiences and the presence of outdated data that can degrade the performance. 
Recently, many algorithms attempt to mitigate these issues with a variety of unique strategies. 
For example, the combined experience replay (CER) is also a sampling strategy that samples fresh experiences using the latest data \cite{CER}.
Alternative replay memory has also been proposed to replay recent experiences without explicit prioritization \cite{remember_forget}. 
There is also an attempt to solve the problem of sampling efficiency in multi-goal tasks.
Hindsight experience replay (HER) uses unsuccessful experiences and considers them as successful experiences achieving different goals, increasing the sampling efficiency \cite{HER}.
However, replay memory contains many experiences regardless of their importance, which can make training inefficient.
To address the inefficiency issue, prioritized experience replay (PER) \cite{PER} samples meaningful experience with a prioritized probability that depends on the temporal-difference (TD) error $\delta$, which has been used as a prioritization metric for determining which experiences to replay, which experiences to store, or which features to select \cite{feature_select,surprise_data}.
On the other hand, because PER prioritizes experiences before sampling, the priorities of all experiences in memory should be calculated, which is inefficient.
Also, sampling with a prioritized probability introduces bias because it changes the distribution of the sampled data.
Additional hyperparameter tuning is required for the importance-sampling weight to reduce the bias \cite{important_sampling}.

In this paper, we propose a new method known as the prioritization-based weighted loss function (PBWL) which is more efficient than PER and which is capable of reducing bias without any additional manipulation.
Additionally, it is shown that PBWL is compatible with PER because these two prioritization methods are applied at two different stages.
The proposed method can improve the performance of the various off-policy RL methods that use replay memory.
The main contributions of this paper are as follows.
\begin{enumerate}
    \item Prioritizing experiences without changing the sampling strategies
    \item Showing that the proposed method is compatible with PER
\end{enumerate}


\section{Background}
\subsection{Temporal Difference Off-policy Reinforcement Learning}
TD prediction, a widely used method in off-policy RL, evaluates policies and uses experiences to update an evaluation network, called a critic network.
Replay memory has played a significant role in the off-policy RL due to the necessity of experiences in off-policy RL.
In particular, replay memory breaks the correlation between the experiences by letting all experiences be uniformly sampled.
There are also many variants when using replay memory. These include which experiences to store, which experiences to replay, and which experiences to forget \cite{remember_forget}.
In this paper, several off-policy types of RL, in this case the deep Q-network (DQN), the deep deterministic policy gradient (DDPG), and the soft actor-critic (SAC), all of which use replay memory \cite{atrai,ddpg,sac}, are tested with and without PBWL.
\subsection{Prioritization}
Prioritization is a technique that is widely used for replay memory \cite{prioritized_sweeping}.
Recently, certain existing sampling methods, which are also prioritization methods, allow more meaningful experiences to be sampled. 
For example, the technique known as PER samples more surprising experiences \cite{PER}, and the magnitude of the TD error can indicate a measure of how surprising the experience is.
In PER, all experiences in replay memory have a priority that is computed using the magnitude of the TD error, and this priority is the probability of being sampled, which is proportional to the TD error.

\section{Proposed Method}
In this section, a novel prioritizing strategy, PBWL, is described.
In earlier studies of PER, each experience is prioritized by the probability of being sampled, which is calculated by the relative magnitude of the TD errors of the experiences in the entire replay memory.
Hence, priority levels can be measured by making relative comparisons with other experiences in replay memory.
PBWL measures the priority of each experience based on the distribution of the TD errors of sampled experiences (DTSE).
It is assumed that the size of the mini-batch is large enough such that DTSE can follow the distribution of TD errors of the entire experience in replay memory.
PBWL samples experiences uniformly and determines which experiences to learn more or learn less, while PER determines which experiences to sample.
\begin{figure*}[t]
    \centering
    \includegraphics[scale=0.6]{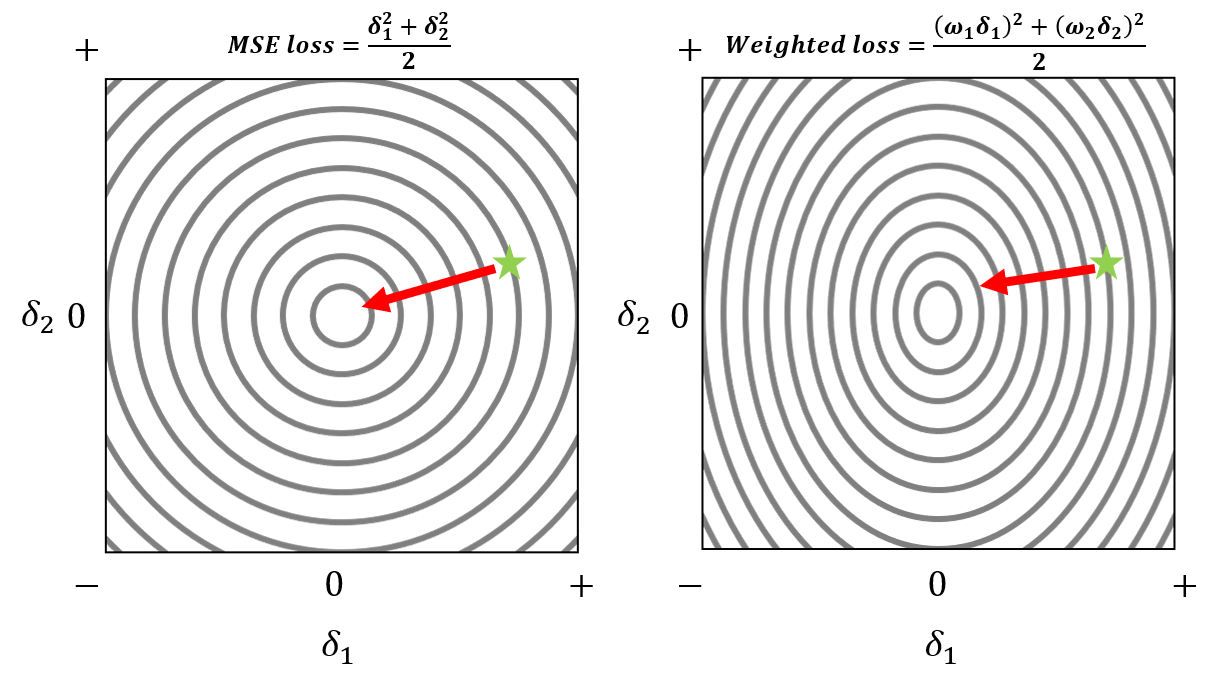}
    \caption{Loss function for two TD errors (batch size: 2). Weighting factors multiplied by TD errors distort the loss function and consequently change the gradient. (a) MSE loss function. (b) Weighted loss function.}
    \label{loss}
\end{figure*}
\begin{figure}[t]
    \includegraphics[scale=0.33]{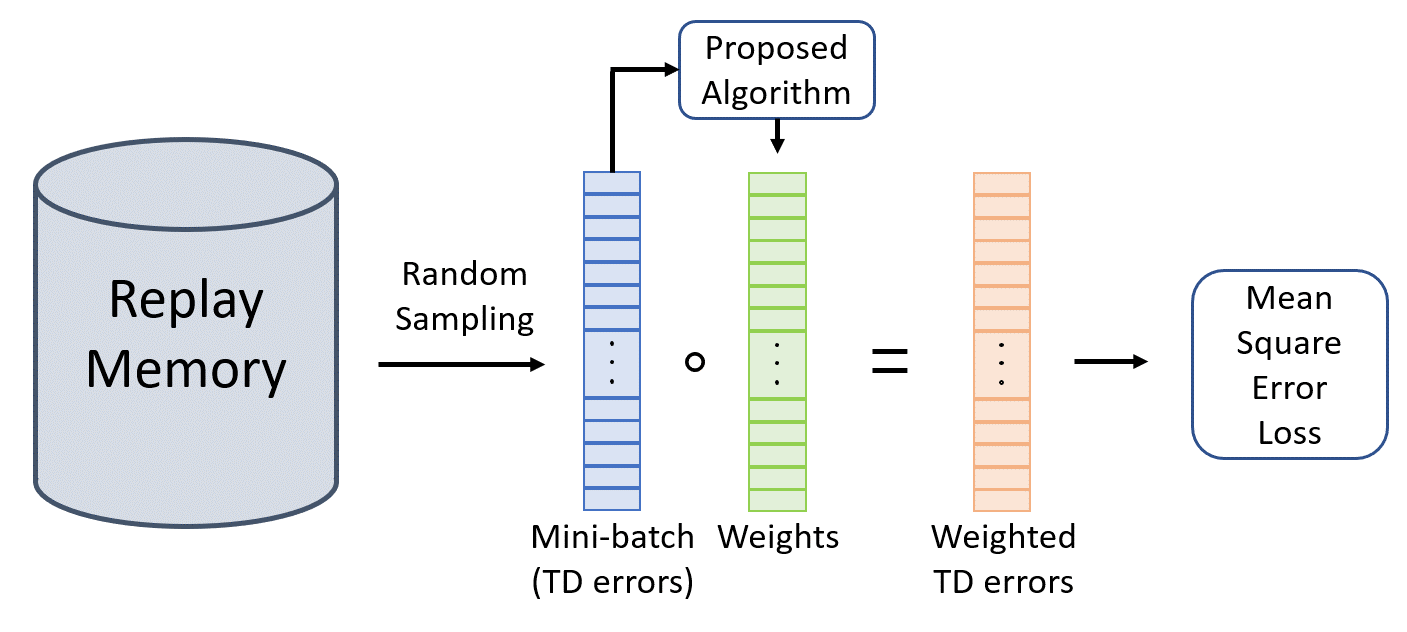}
    \caption{Overall structure of the proposed method. TD errors of the randomly sampled mini-batch are element-wise multiplied with weights of the same size to calculate the MSE loss.}
    \label{algorithm_structure}
\end{figure}
In various machine learning fields, several prior studies have attempted to improve the performance of existing algorithms by introducing a weighting factor \cite{weight_factor1,weight_factor2,weight_factor3,weight_factor4,weight_factor5,weight_factor6}.
The likelihood-free importance weighting method re-weights experiences based on their likelihood under the stationary distribution of the current policy by using a weighting factor \cite{weighting_to_loss}.
In this paper, PBWL, also introduces a weighting factor into each sample, leading to prioritization.
More specifically, prioritization can be achieved by multiplying the different values by the error of each, as shown in Figure \ref{algorithm_structure}.
In other words, the proposed method prioritizes each experience in a mini-batch by multiplying the \begin{math}j_{th}\end{math} TD error \begin{math}\delta_{j}\end{math} of the experience by the \begin{math}j_{th}\end{math} weight \begin{math}\omega_{j}\end{math}.
We focus on the mean square error (MSE) loss function in this paper given that the MSE is used as a loss function in many RL algorithms.
\begin{equation}
\mathcal{L}_{MSE}=\cfrac{1}{N}\sum_{j=1}^N\delta_{j}^2\label{MSE}
\end{equation}
\begin{equation}
\mathcal{L}_{W}=\cfrac{1}{N}\sum_{j=1}^N(\omega_{j}\delta_{j})^2\label{W}
\end{equation}
N is the mini-batch size and \begin{math}\mathcal{L}_{MSE}\end{math} and \begin{math}\mathcal{L}_{W}\end{math} are, respectively, the MSE loss and the weighted loss.
The higher the weight \begin{math}\omega_{j}\end{math} is, the more preferentially TD error \begin{math}\delta_{j}\end{math} is optimized because this weighted loss function in \eqref{W} consists of the square sum of the weighted TD error.
The weight \begin{math}\omega\end{math} distorts the loss function, a function of \begin{math}\delta\end{math}, to be advantageous for optimization for a specific TD error. 
In other words, distortion of the loss function changes the direction of the gradient, as shown in Figure \ref{loss}, which can be mathematically shown in \eqref{gradient1}, \eqref{gradient2} and \eqref{magnitude}.
\begin{equation}
\boldsymbol{\theta}\leftarrow\boldsymbol{\theta}-\alpha\nabla_{\boldsymbol{\theta}}\mathcal{L}(\boldsymbol{\delta};\boldsymbol{\theta})\label{gd}
\end{equation}
\begin{equation}
\nabla_{\boldsymbol{\theta}}\mathcal{L}=
\begin{bmatrix} \cfrac{\partial\mathcal{L}}{\partial\theta_{1}} \\ \cfrac{\partial\mathcal{L}}{\partial\theta_{2}} \\ \vdots \\ \cfrac{\partial\mathcal{L}}{\partial\theta_{K}} \end{bmatrix}=
\begin{bmatrix} 
\cfrac{\partial\mathcal{L}}{\partial\delta_{1}}\cdot\cfrac{\partial\delta_{1}}{\partial\theta_{1}}+\cdots+\cfrac{\partial\mathcal{L}}{\partial\delta_{N}}\cdot\cfrac{\partial\delta_{N}}{\partial\theta_{1}} \\ 
\cfrac{\partial\mathcal{L}}{\partial\delta_{1}}\cdot\cfrac{\partial\delta_{1}}{\partial\theta_{2}}+\cdots+\cfrac{\partial\mathcal{L}}{\partial\delta_{N}}\cdot\cfrac{\partial\delta_{N}}{\partial\theta_{2}} \\ 
\vdots \\ 
\cfrac{\partial\mathcal{L}}{\partial\delta_{1}}\cdot\cfrac{\partial\delta_{1}}{\partial\theta_{K}}+\cdots+\cfrac{\partial\mathcal{L}}{\partial\delta_{N}}\cdot\cfrac{\partial\delta_{N}}{\partial\theta_{K}} 
\end{bmatrix}\label{gradient1}
\end{equation}
\begin{equation}
\nabla_{\boldsymbol{\theta}}\mathcal{L}=\cfrac{\partial\mathcal{L}}{\partial\delta_{1}}\cdot\nabla_{\boldsymbol{\theta}}\delta_{1}+\cdots+\cfrac{\partial\mathcal{L}}{\partial\delta_{N}}\cdot\nabla_{\boldsymbol{\theta}}\delta_{N}\label{gradient2}
\end{equation}
\begin{equation}
\cfrac{\partial\mathcal{L}}{\partial\delta_{j}}=
\left
\{
\begin{array}{rcl}
\cfrac{2\delta_{j}}{N} & \mbox{for} & \mathcal{L}=\mathcal{L}_{MSE} \\ 
\cfrac{2\delta_{j}\omega_{j}}{N} & \mbox{for} & \mathcal{L}=\mathcal{L}_{W}
\end{array}\right.
\label{magnitude}.
\end{equation}

Equation \eqref{gd} presents the generally used gradient descent method to update 
\begin{math}
[\theta_{1},\theta_{2},\cdots,\theta_{K}]^{\mathsf{T}}=\boldsymbol{\theta}\in\mathbb{R}^{K}
\end{math},
a column vector of K parameters, indicating \begin{math}[\cdot]^{\mathsf{T}}\end{math} the transpose of the vector \begin{math}[\cdot]\end{math}.
The gradient of loss function is depicted here as \eqref{gradient1}, which can be also represented as \eqref{gradient2}.
This \eqref{gradient2} indicates that the gradient of loss function consists of the sum of vectors that are aligned with the direction of optimization of each TD error. 
Equation \eqref{magnitude} is a partial derivative of the loss function for TD error \begin{math}\delta_{j}\end{math} and is multiplied by the vector \begin{math}\nabla_{\boldsymbol{\theta}}\delta_{j}\end{math} so that it influences the magnitude of the vector.
\begin{equation}
\omega_{j}=\omega(\delta_{j}|\delta_{1},\delta_{2},\cdots,\delta_{N})=\omega(\delta_{j}|\boldsymbol{\delta})\label{omega}
\end{equation}
The \begin{math}j_{th}\end{math} weight \begin{math}\omega_{j}\end{math} is computed as \eqref{omega} using the magnitude of TD errors \begin{math}\boldsymbol{\delta}\in\mathbb{R}^{N}\end{math}, the vector of TD errors in the N mini-batch, but it is not entirely proportional to the \begin{math}j_{th}\end{math} TD error \begin{math}\delta_{j}\end{math}.
Our hypothesis is that such a well-trained model will aptly fit most of the samples.
In other words, the main target to learn is the range in which the magnitudes of TD errors are concentrated. 
Moreover, the range in which the magnitudes of TD errors are high is also meaningful to learn, as a high level of TD error indicates a measure of how surprising the experience is \cite{PER}.
Therefore, we propose a new criterion to prioritize experiences in this paper. 
An experience of which the TD error \begin{math}\delta_{j}\end{math} is close to the middle of the DTSE is prioritized over most experiences.
Furthermore, comparing the experiences of which the TD errors are far from the middle of the DTSE, thosefar from the middle toward the positive direction are prioritized over others far from the middle toward the negative direction. 
The properties of the weighting factor are expressed as \eqref{weighting_factor_property}.\\
\begin{equation}
\begin{cases}
\omega(\delta_{x}|\boldsymbol{\delta})>\omega(\delta_{x}+\epsilon|\boldsymbol{\delta})&\delta_{x,n}>0\\
\omega(\delta_{x}|\boldsymbol{\delta})<\omega(\delta_{x}+\epsilon|\boldsymbol{\delta})&\delta_{x,n}<0\\
\omega(\delta_{y}|\boldsymbol{\delta})<\omega(\delta_{x}|\boldsymbol{\delta})&\delta_{y,n}=-\delta_{x,n},\delta_{y}<\delta_{x}\label{weighting_factor_property}
\end{cases}
\end{equation}
where $\delta_{,n}$ is the normalized TD error, which is described in the following section.
\begin{figure}[t]
    \includegraphics[scale=0.53]{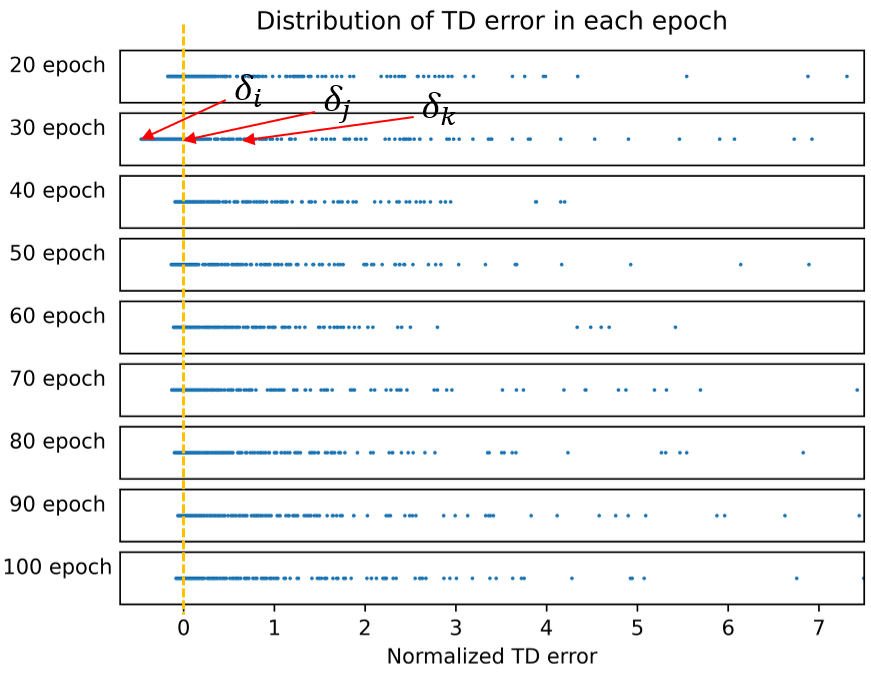}
    \caption{Distribution of the TD error during training in the FetchPush-v1 environment. $\delta_{j}$ is in the middle of DTSE, and $\delta_{i}$ and $\delta_{k}$ differ from the middle of DTSE by the same amount of magnitude. This figure shows that the distribution of the TD error is negatively skewed.}
    \label{distribution_of_TD_error}
\end{figure}
Taking an absolute value for every TD error comes first because the priority depends on the magnitude of the TD error.
It is important to note that the process described in the next section sets the weight of each experience rather than changing the TD error itself. 
The proposed method consists of five processes, and each specific process is presented in the following sections.

\subsection{Combined Normalization}
The range of the TD errors sampled can differ depending on the environment with which the agent interacts.
Normalization with the middle of DTSE and the standard deviation of the TD error can be done generally to apply the PBWL method.
However, the mean of the TD error cannot suitably represent the middle of the DTSE when there are many outliers.
Moreover, as the agent learns, there is a likelihood that experiences that are unfamiliar may be stored in replay memory. 
The DTSE can be negatively skewed given that an experience not familiar to the agent is likely to have a high TD error \cite{skewed1,skewed2,skewed3,skewed4}, as shown in Figure \ref{distribution_of_TD_error}.
Normalization with the median of TD errors represents the middle of the DTSE well due to the negatively skewed shape of the DTSE.
On the other hand, we assume that normalization with the mean can conservatively train the network in a situation where the DTSE is positively skewed.
This means that prioritizing the experiences with a relatively low magnitude of TD error mitigates the overheating of the training process by preventing any dramatic changes in the neural network.
In this paper, we use a variant of this type of normalization referred to as combined normalization. It is defined as follows:
\begin{equation}
\delta(j)_{n}=\cfrac{\lvert\delta(j)\rvert-\min(\mu(\lvert\boldsymbol{\delta}\rvert), m(\lvert\boldsymbol{\delta}\rvert))}{\sigma(\lvert\boldsymbol{\delta}\rvert)}\label{normalize}
\end{equation}
where $\mu(\lvert\boldsymbol{\delta}\rvert)$, $m(\lvert\boldsymbol{\delta}\rvert)$, and $\sigma(\lvert\boldsymbol{\delta}\rvert)$ are respectively the mean, median, and standard deviation of the absolute value of all TD errors  in the mini-batch and where $\delta(j)$ and $\delta(j)_{n}$ are the $j_{th}$ TD error and the normalized TD error, respectively.

\subsection{Positive preferential function}
With combined normalization, TD errors that are close to the middle of the DTSE have normalized TD errors approaching zero.
In other words, the experiences we want to prioritize as higher have normalized TD errors approaching zero. 
However, in the context of a comparison between larger (positive) and smaller (negative) normalized TD errors which are not close to zero, the priority cannot be distinguished.
The modifying strategy used in this paper can be mathematically shown to ensure that a positive normalized TD error has a property similar  to that of a normalized TD error which is close to zero, as follows:
\begin{equation}
\delta(j)_{m}=\delta(j)_{n}\times\cfrac{1}{\max(\delta(j)_{n},0)+1}\label{modify}
\end{equation}
where $\delta(i)_{m}$ is the modified normalized $i_{th}$ TD error. We refer to this modifying strategy as a positive preferential function in this paper.

\subsection{Gaussian function}
Gaussian functions are used as a filter to reduce outliers by assigning less weight to values further from zero \cite{gaussian}. 
The use of a Gaussian function with a zero mean can ensure that experiences with modified normalized TD errors close to zero have high priority levels and can reduce the effects of outliers. This can be mathematically described as
\begin{equation}
\delta(j)_{g}=\cfrac{1}{\sqrt{2\pi}\times\sigma(\boldsymbol{\delta_{m}})}\exp\bigg(-\cfrac{1}{2}\Big(\cfrac{\delta(j)_{m}}{\sigma(\boldsymbol{\delta_{m}})}\Big)^{2}\bigg)\label{gaussian}
\end{equation}
where $\delta(j)_{g}$ is the raw priority of the $j_{th}$ experience.
The Gaussian function also helps to reduce the effects of outliers.

\subsection{Softmax function}
The experiences for which we want to assign high priority levels have a high raw priority level $\delta(j)_{g}$ but the differences among them make their use excessive. 
Furthermore, the raw priority represents differences in the priority level but has no meaning in itself.
In other words, only differences between the values are considered. 
The softmax function, commonly used in deep neural networks, can be used to convert values into action probabilities in RL. 
However, in this paper, this function is used to prevent dead experiences from being introduced and to rescale the raw priorities. 
The softmax function used in this paper is expressed as follows:
\begin{equation}
p_{j}=\sigma_{s}(\delta(j)_{g})=\cfrac{\exp(\delta(j)_{g})}{\sum_{i=1}\exp(\delta(i)_{g})}\label{softmax}
\end{equation}
where $\sigma_{s}(\cdot)$ is the softmax operation and $p_{j}$ is the $j_{th}$ weighting factor (WF) for multiplication to the $j_{th}$ experience.
One of the crucial properties of the softmax function is that it is invariant to a constant offset, which means only differences between values are considered.
This can be mathematically expressed as follows:
\begin{align}
\sigma_{s}(z_{k}+A)=&\cfrac{\exp(z_{k}+A)}{\sum_{i=1}\exp(z_{i}+A)}\nonumber\\
=&\cfrac{\exp(z_{k})}{\sum_{i=1}\exp(z_{i})}=\sigma_{s}(z_{k})\label{invariant}
\end{align}
Let $\boldsymbol{X}=[0.2,0.4,0.6]$ be the vector of three raw priorities.
The differences between those raw priorities are 0.2.
The distribution of $\boldsymbol{X}$ can differ, such as $\boldsymbol{X_{1}}=[0.4,0.6,0.8]$ depending on the environment in which the agent interacts or on the mini-batch in replay memory. 
Because the softmax function is invariant to translation, the outputs of $\boldsymbol{X}$ and $\boldsymbol{X_{1}}$ are the same.
Moreover, the softmax function as used here reduces the likelihood of dead experiences occurring when $\delta(j)_{g}\approx0$, preventing the overfitting of certain experiences, given that the lower limit of the $\exp(\delta(j)_{g})$ is 1.
Also, the excessive proportional difference between the WFs can be alleviated by rescaling.

\begin{figure}[t]
    \centering
    \includegraphics[scale=0.43]{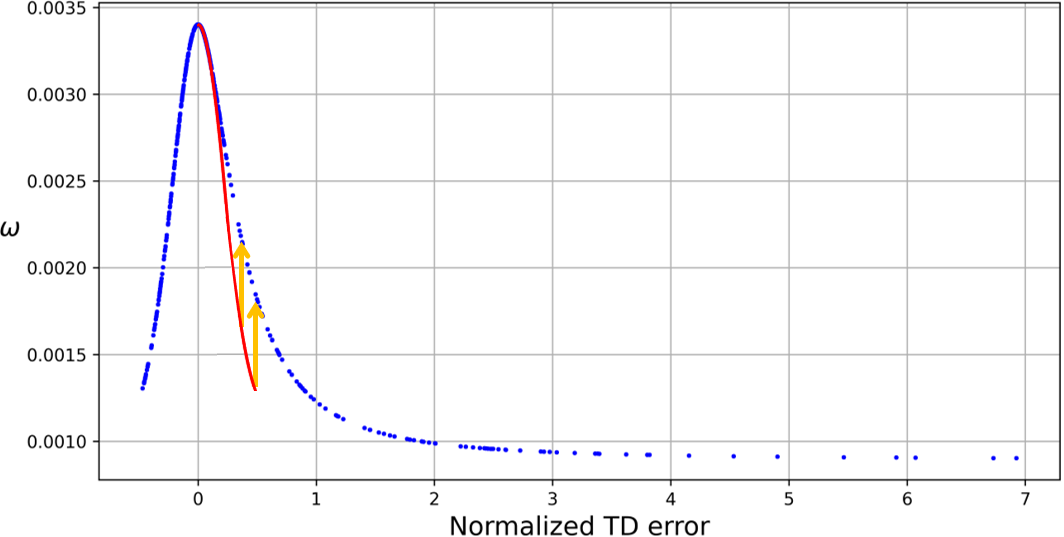}
    \caption{The weighting factor $\omega$ for the normalized TD error. The red line represents the weighting factor when the positive prefer function is not used.}
    \label{procedure}
\end{figure}

\subsection{Compensation for the loss function}
The rescaled priority, WF, indicates how weighted each experience is, and it is multiplied by each TD error.
However, each WF lowers the loss function because the sum of the WFs is 1, as presented in \eqref{unstable_loss}. 
Considering how the parameters in a deep neural network are updated, as described in \eqref{gd}, changes in the loss function will likely deteriorate the stability.
\begin{equation}
\mathcal{L}_{MSE}=\cfrac{1}{N}\sum_{i=1}^{N}\delta_{i}^{2}>\cfrac{1}{N}\sum_{i=1}^{N}(p_{i}\delta_{i})^{2}\label{unstable_loss}
\end{equation}
Due to stability concerns, compensation for changes in the loss function is inevitable.
The weighted loss function can be calculated by multiplying the compensation factor and the WF by the TD error, as follows:
\begin{equation}
\cfrac{1}{N}\sum_{i=1}^{N}\Big(\cfrac{\lVert\boldsymbol{\delta}\rVert_{1}}{\lVert \boldsymbol{p}\circ\boldsymbol{\delta}\rVert_{1}}p_{i}\delta_{i}\Big)^{2}=\cfrac{1}{N}\sum_{i=1}^{N}(\omega_{i}\delta_{i})^{2}=\mathcal{L}_{W}\label{compensated_loss}
\end{equation}
where $\cfrac{\lVert\boldsymbol{\delta}\rVert_{1}}{\lVert \boldsymbol{p}\circ\boldsymbol{\delta}\rVert_{1}}$  is the compensation factor, and $\circ$ and $\lVert\cdot\rVert_{1}$ are, respectively, the element-wise product operation and the L1-norm operation.
The L1-norm is used in this paper instead of the L2-norm because the former is more robust to outliers while the latter is likely to favor outliers.

\begin{algorithm}[tb]
    \caption{Proposed method: Prioritization-Based Weighted Loss}
    \label{PBWL}
    \textbf{Given}: mini-batch size $N$, the sampled TD errors $\boldsymbol{\delta}$, the normalized TD errors $\boldsymbol{\delta_{n}}$, the modified normalized TD errors $\boldsymbol{\delta_{m}}$, the raw priorities $\boldsymbol{\delta_{g}}$, the weighting factors $\boldsymbol{p}$, and the compensated weighting factors $\boldsymbol{\omega}$.
    
    $(\boldsymbol{\delta},\boldsymbol{\delta_{n}},\boldsymbol{\delta_{m}},\boldsymbol{\delta_{g}},\boldsymbol{p},\boldsymbol{\omega})\in\mathbb{R}^{N}$\\
    \textbf{Input}: TD errors $\boldsymbol{\delta}=[\delta_{1},\delta_{2},\delta_{3},\cdots,\delta_{N}]$ in a random mini-batch of $N$ experiences\\
    \textbf{Output}: Compensated weighting factor $\boldsymbol{\omega}$
    
    \begin{algorithmic}[1] 
        \STATE Normalize the absolute values of TD errors with combined normalization.\\
        $\boldsymbol{\delta_{n}}\leftarrow\cfrac{\lvert\boldsymbol{\delta}\rvert-\min(\mu(\lvert\boldsymbol{\delta}\rvert),m(\lvert\boldsymbol{\delta}\rvert))}{\sigma(\lvert\boldsymbol{\delta}\rvert)}$
        \STATE Modify the positive normalized TD errors close to zero.\\
        $\boldsymbol{\delta_{m}}\leftarrow\boldsymbol{\delta_{n}}\times\cfrac{1}{\max(\boldsymbol{\delta_{n}},0)+1}$
        \STATE Compute the raw priority $\boldsymbol{\delta_{g}}$.\\
        $\boldsymbol{\delta_{g}}\leftarrow\cfrac{1}{\sqrt{2\pi}\times\sigma(\boldsymbol{\delta_{m}})}\exp\bigg(-\cfrac{1}{2}\Big(\cfrac{\boldsymbol{\delta_{m}}}{\sigma(\boldsymbol{\delta_{m}})}\Big)^{2}\bigg)$
        \STATE Compute the weighting factor $\boldsymbol{p}$.\\
        $\boldsymbol{p}\leftarrow\cfrac{\exp(\boldsymbol{\delta_{g}})}{\sum_{i=1}\exp(\delta(i)_{g})}$
        \STATE Compute the compensated weighting factor $\boldsymbol{\omega}$.\\
        $\boldsymbol{\omega}\leftarrow\cfrac{\lVert\boldsymbol{\delta}\rVert_{1}}{\lVert \boldsymbol{p}\circ\boldsymbol{\delta}\rVert_{1}}\times \boldsymbol{p}$
        \STATE \textbf{return} $\boldsymbol{\omega}=[\omega_{1},\omega_{2},\omega_{3},\cdots,\omega_{N}]$
    \end{algorithmic}
\end{algorithm}
\begin{figure}[t]
    \includegraphics[scale=0.85]{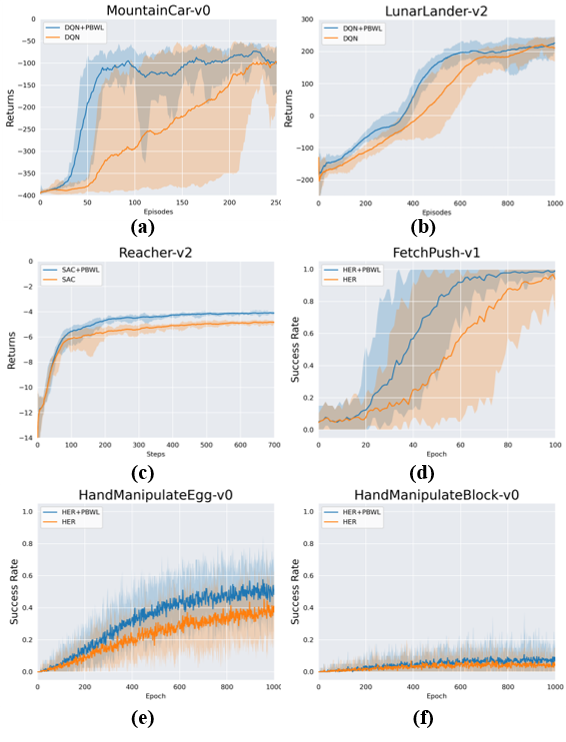}
    \caption{Learning curves for the suite of OpenAI Gym control tasks. Plots (a) and (b) represent the learning curves for the discrete control tasks, and plots (c), (d), (e) and (f) represent the learning curves for the continuous control tasks.}
    \label{result}
\end{figure}
\section{Experiments}
The benefit of the proposed method, PBWL, is evaluated in the OpenAI Gym environment \cite{openaigym}. 
PBWL is combined with TD off-policy RL algorithms, specifically DQN, DDPG, and SAC, for a comparison with the PBWL algorithm and the baselines.
All of the experiments in Figure \ref{result}  are repeated with ten random seeds. Accordingly, ten curves are obtained and the mean of the values of the ten curves is selected and presented as the result.
Experiments are executed in two different discrete control environments and four different continuous control environments.
The RL algorithms used in these experiments as a baseline are representative algorithms in these environments.

\subsection{Discrete control}
The proposed method, PBWL, is tested with DQN, which is widely used in discrete control applications, in the MountainCar-v0 and LunarLander-v2 tasks.
The goals of MountainCar-v0 and LunarLander-v2 are, respectively, to drive up the mountain on the right by accelerating the car to the right or left and to direct the agent to a landing pad as softly and fuel-efficiently as possible.
Originally, in the MountainCar-v0 case, a reward of -1 is given for every step until an episode, consisting of 200 steps, ends. However, a reward-shaping technique is used to make the problem easier to learn \cite{reward_shaping}.
In this paper, a reward of -2 is given for every step, and a reward of +100 for the goal and the normalized mechanical energy, which is the sum of the potential energy and kinetic energy, is also given.

Figures \ref{result} (a) and (b) show the results of the MountainCar-v0 and LunarLander-v2 tasks. 
The training is, respectively, performed for 250 and 1000 episodes. 
For each training episode, the returns are defined as the sum of the reward during an episode. 
To reduce granularity at each episode, the moving average of the returns across, respectively, 10 and 100 episodes is taken.
We find that DQN with PBWL outperforms the baseline. 
The training performance of the proposed method is, respectively, better than that of the baseline over the ranges of episode indexes of 60-250 and 600-1000.

\subsection{Continuous control}
Experiments are executed with the MuJoCo physics engine \cite{mujoco} and the robotic environment in the OpenAI Gym for continuous control tasks. 
The environments of the experiments considered in this paper are Reacher-v2, FetchPush-v1, HandManipulateEgg-v0, and HandManipulateBlock-v0. 
For Reacher-v2, PBWL is tested with SAC, a state-of-the-art algorithm in this environment \cite{sac_test}. 
The training is performed for 70000 steps and, after every 100 steps, ten evaluations are executed.
In the last evaluation, the SAC with PBWL achieve a return value close to -4, while the baseline SAC achieve a return value close to -5.
We find the SAC with PBWL improves over the baseline, with a performance gain of approximately 11\% as presented in Figure \ref{result} (c).
DDPG with HER, which is a widely used TD off-policy algorithm for multi-goal continuous control, is tested as a baseline for the other three environments: FetchPush-v1, HandManipulateEgg-v0, and HandManipulateBlock-v0. 
Figures \ref{result} (d), (e), and (f) present, respectively, the results of the FetchPush-v1, HandManipulateEgg-v0, and HandManipulateEgg-v0 tasks.
We find that the baseline with PBWL matches or outperforms the baseline in the presented tasks.
The training performance of the proposed method is better than that of the baseline algorithm over the epoch index range of 60-100 in FetchPush-v1. 
For the HandManipulateEgg-v0 and HandManipulateBlock-v0 environments, PBWL increases and marginally increases the mean success rate at the epoch index of 1000 by approximately 10\% and 3\%, respectively.
The training is, respectively, performed for 100, 1000, and 1000 epochs. 
In one epoch, 50 episode cycles and 20 evaluations are executed sequentially.
\begin{figure}[t]
    \includegraphics[scale=1]{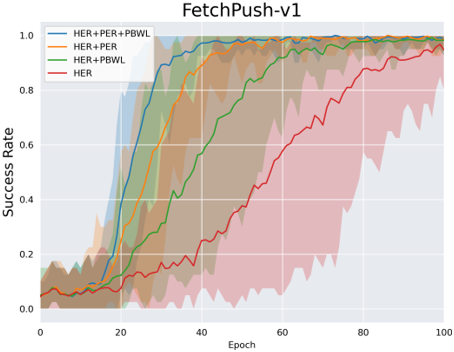}
    \caption{Learning curves of FetchPush-v1 for compatibility with PER.}
    \label{compatibility_ablation}
\end{figure}
\begin{table*}[t]
    \centering
    \begin{tabular}{c|c|c|cccc}
        \toprule
        Normalization & Softmax & Norm & 25 epochs & 50 epochs & 75 epochs & 100 epochs \\
        \midrule
        \multirow{4}*{Combined}  & \multirow{2}*{On}   & L1 & $\textbf{0.20}\pm\textbf{0.03}$ & $\textbf{0.65}\pm\textbf{0.02}$ & $\textbf{0.96}\pm\textbf{0.02}$ & $\textbf{0.98}\pm\textbf{0.00}$\\
                                 &                     & L2 & 0.11$\pm$0.02 & 0.20$\pm$0.01 & 0.50$\pm$ 0.07 & 0.74$\pm$0.02\\
                                 & \multirow{2}*{Off}  & L1 & 0.10$\pm$0.03 & 0.07$\pm$0.02 & 0.06$\pm$0.03 & 0.06$\pm$0.02\\
                                 &                     & L2 & 0.10$\pm$0.03 & 0.07$\pm$0.02 & 0.06$\pm$0.03 & 0.06$\pm$0.02\\
        \multirow{4}*{Mean}      & \multirow{2}*{On}   & L1 & 0.12$\pm$0.04 & 0.12$\pm$0.04 & 0.17$\pm$0.00 & 0.17$\pm$0.02\\
                                 &                     & L2 & 0.11$\pm$0.03 & 0.07$\pm$0.03 & 0.08$\pm$0.02 & 0.07$\pm$0.04\\
                                 & \multirow{2}*{Off}  & L1 & 0.10$\pm$0.03 & 0.07$\pm$0.02 & 0.07$\pm$0.02 & 0.05$\pm$0.03\\
                                 &                     & L2 & 0.10$\pm$0.03 & 0.07$\pm$0.02 & 0.06$\pm$0.02 & 0.06$\pm$0.03\\
        \multirow{4}*{Median}    & \multirow{2}*{On}   & L1 & $\textbf{0.20}\pm\textbf{0.03}$ & $\textbf{0.65}\pm\textbf{0.02}$ & $\textbf{0.96}\pm\textbf{0.02}$ & $\textbf{0.98}\pm\textbf{0.00}$\\
                                 &                     & L2 & 0.12$\pm$0.02 & 0.21$\pm$0.02 & 0.47$\pm$0.06 & 0.68$\pm$0.02\\
                                 & \multirow{2}*{Off}  & L1 & 0.10$\pm$0.03 & 0.07$\pm$0.02 & 0.06$\pm$0.03 & 0.06$\pm$0.02\\
                                 &                     & L2 & 0.10$\pm$0.03 & 0.07$\pm$0.02 & 0.07$\pm$0.02	& 0.06$\pm$0.02\\
        \bottomrule
    \end{tabular}
    \caption{Ablation studies in FetchPush-v1}
    \label{ablation_table}
\end{table*}
\subsection{Compatibility with PER}
In this section, we present the compatibility of PBWL with PER, a widely used prioritization method. 
The proposed method, PBWL, prioritizes experiences to learn more after sampling, while PER prioritizes experiences to sample more surprising experiences before sampling.
This indicates that these two algorithms can be applied in different steps. 
We assume that PBWL can be used with PER, improving the performance of the algorithm.
The test is executed in FetchPush-v1 and is repeated with ten random seeds, after which the mean of the values of the ten curves is selected and presented as the result.
Figure \ref{compatibility_ablation} shows that PBWL can be applied with PER. 
In this test, DDPG with HER is tested as a baseline. 
The training performance of PBWL with PER and the baseline is better than that of the others over the epoch index range of 15-100.

\subsection{Ablation study}
The benefit of the proposed method, PBWL, is evaluated in the OpenAI Gym environment. 
The proposed method consists of five components, of which three contribute to the stability of the algorithm: Combined normalization, the softmax function, and compensation with L1-norm.
Combined normalization, the first component, is normalization robust to outliers.
A combination of the mean and median is used to normalize so as suitably to represent the middle of the distribution of the experiences.
Furthermore, the softmax function can alleviate any excessive imbalance between the values so as to prevent some experiences from assignment of an overly priority level. 
Finally, compensation with L1-norm can also augment the stability.
L1-norm is used to ignore the effects of outliers on the loss function, as some experiences with high error levels can greatly increase the loss function. 
We conduct ablation studies to investigate the effects of the three aforementioned components in an effort to understand the contributions of these three components to the proposed PBWL algorithm.
The results of the ablation studies are presented in Table \ref{ablation_table}.
The results show that the median normalization and combined normalization strategies perform similarly, whereas mean normalization does not work at all.
It indicates that the mean can’t represent the middle of the distribution well.
In particular, these two normalization methods (excluding mean normalization) show no differences when the softmax function and L1-norm are used simultaneously.
However, there is a difference between the combined normalization method and the mean normalization method when the softmax function and L2 norm are used at the same time.
This shows that the use of combined normalization instead of median normalization marginally increases the performance. 
The use of the softmax function has also a strong influence on the training, as shown in Figure \ref{compatibility_ablation} (b) and Table \ref{ablation_table}.
The softmax function prevented the sample from becoming excessively unbalanced, which means that this function prevents the neural network from overfitting specific samples. 
Among them, the least impact on performance is the ablation of the compensation with L1-norm.
Whether or not to use the combined or the median normalization strategies and whether or not to use the softmax function will have a significant impact on the performance.
On the other hand, whether or not to compensate with L1-norm instead of L2 norm is helpful when training, but it has no significant impact compared to the normalization method and the softmax function.

\section{Conclusion}
Training agents via off-policy deep RL requires many experiences, which can be achieved with ER. 
Sampling random experiences from memory can hinder learning due to different levels of importance among experiences. 
In other words, some experiences are useful while others are useless. 
In this paper, we introduce a new method to prioritize experiences based on DTSE without a sampling strategy.
The proposed method, PBWL, matches or outperforms the baseline algorithms while also reducing bias given that all experiences are uniformly sampled from memory. 
Furthermore, we show that PBWL can be used with PER, a widely used sampling strategy.
In this paper, the results indicate that weighting each sample in a mini-batch can improve the performance of the algorithm.


\begin{table}[t]
\begin{tabular}{cccc}\toprule
\multirow{2}{*}{\begin{tabular}[c]{@{}c@{}}\multirow{1.5}*{Batch}\\ \multirow{1.5}*{size}\end{tabular}} & \multicolumn{2}{c}{Average convergence epochs} & \multirow{2}{*}{\begin{tabular}[c]{@{}c@{}}\multirow{1.5}*{Reduction}\\ \multirow{1.5}*{Rate (\%)}\end{tabular}} \\\cmidrule(lr){2-3}
     & w/o PBWL & w/ PBWL      &      \\\midrule
128  & 82.5     & not converge & -     \\
256  & 53.2     & 34.3         & 35.5 \% \\
512  & 39.3     & 23.3         & 40.7 \% \\\bottomrule
\end{tabular}
\caption{Ablation studies for batch size in FetchPush-v1}
\label{ablation_batch}
\end{table}

\bibliographystyle{named}
\bibliography{ijcai22}

\begin{thebibliography}{}

\bibitem[\protect\citeauthoryear{Anand \bgroup \em et al.\egroup
  }{2010}]{weight_factor1}
Ashish Anand, Ganesan Pugalenthi, Gary~B Fogel, and PN~Suganthan.
\newblock An approach for classification of highly imbalanced data using
  weighting and undersampling.
\newblock {\em Amino acids}, 39(5):1385--1391, 2010.

\bibitem[\protect\citeauthoryear{Andrychowicz \bgroup \em et al.\egroup
  }{2017}]{HER}
Marcin Andrychowicz, Filip Wolski, Alex Ray, Jonas Schneider, Rachel Fong,
  Peter Welinder, Bob McGrew, Josh Tobin, OpenAI Pieter~Abbeel, and Wojciech
  Zaremba.
\newblock Hindsight experience replay.
\newblock {\em Advances in neural information processing systems}, 30, 2017.

\bibitem[\protect\citeauthoryear{Au and Beck}{2003}]{important_sampling}
Siu-Kui Au and JL~Beck.
\newblock Important sampling in high dimensions.
\newblock {\em Structural safety}, 25(2):139--163, 2003.

\bibitem[\protect\citeauthoryear{Brockman \bgroup \em et al.\egroup
  }{2016}]{openaigym}
Greg Brockman, Vicki Cheung, Ludwig Pettersson, Jonas Schneider, John Schulman,
  Jie Tang, and Wojciech Zaremba.
\newblock Openai gym.
\newblock {\em arXiv preprint arXiv:1606.01540}, 2016.

\bibitem[\protect\citeauthoryear{Chan \bgroup \em et al.\egroup
  }{2019}]{sac_test}
Stephanie~CY Chan, Samuel Fishman, John Canny, Anoop Korattikara, and Sergio
  Guadarrama.
\newblock Measuring the reliability of reinforcement learning algorithms.
\newblock {\em arXiv preprint arXiv:1912.05663}, 2019.

\bibitem[\protect\citeauthoryear{Chan \bgroup \em et al.\egroup
  }{2022}]{skewed4}
Stephanie~CY Chan, Andrew~K Lampinen, Pierre~H Richemond, and Felix Hill.
\newblock Zipfian environments for reinforcement learning.
\newblock {\em arXiv preprint arXiv:2203.08222}, 2022.

\bibitem[\protect\citeauthoryear{Cho \bgroup \em et al.\egroup
  }{2022}]{predicting_accident}
Injoon Cho, Praveen~Kumar Rajendran, Taeyoung Kim, and Dongsoo Har.
\newblock Reinforcement learning for predicting traffic accidents.
\newblock {\em arXiv preprint arXiv:2212.04677}, 2022.

\bibitem[\protect\citeauthoryear{Cui \bgroup \em et al.\egroup
  }{2019}]{weight_factor3}
Yin Cui, Menglin Jia, Tsung-Yi Lin, Yang Song, and Serge Belongie.
\newblock Class-balanced loss based on effective number of samples.
\newblock In {\em Proceedings of the IEEE/CVF conference on computer vision and
  pattern recognition}, pages 9268--9277, 2019.

\bibitem[\protect\citeauthoryear{Geramifard \bgroup \em et al.\egroup
  }{2011}]{feature_select}
Alborz Geramifard, Finale Doshi, Josh Redding, Nicholas Roy, and Jonathan~P
  How.
\newblock Online discovery of feature dependencies.
\newblock In {\em ICML}, 2011.

\bibitem[\protect\citeauthoryear{Guo \bgroup \em et al.\egroup
  }{2022}]{weight_factor6}
Dandan Guo, Zhuo Li, Meixi Zheng, He~Zhao, Mingyuan Zhou, and Hongyuan Zha.
\newblock Learning to re-weight examples with optimal transport for imbalanced
  classification.
\newblock {\em arXiv preprint arXiv:2208.02951}, 2022.

\bibitem[\protect\citeauthoryear{Haarnoja \bgroup \em et al.\egroup
  }{2018}]{sac}
Tuomas Haarnoja, Aurick Zhou, Pieter Abbeel, and Sergey Levine.
\newblock Soft actor-critic: Off-policy maximum entropy deep reinforcement
  learning with a stochastic actor.
\newblock In {\em International conference on machine learning}, pages
  1861--1870. PMLR, 2018.

\bibitem[\protect\citeauthoryear{Hong \bgroup \em et al.\egroup }{2021}]{lab6}
Chansol Hong, Inbae Jeong, Luiz~Felipe Vecchietti, Dongsoo Har, and Jong-Hwan
  Kim.
\newblock Ai world cup: Robot-soccer-based competitions.
\newblock {\em IEEE Transactions on Games}, 13(4):330--341, 2021.

\bibitem[\protect\citeauthoryear{Jaderberg \bgroup \em et al.\egroup
  }{2016}]{skewed1}
Max Jaderberg, Volodymyr Mnih, Wojciech~Marian Czarnecki, Tom Schaul, Joel~Z
  Leibo, David Silver, and Koray Kavukcuoglu.
\newblock Reinforcement learning with unsupervised auxiliary tasks.
\newblock {\em arXiv preprint arXiv:1611.05397}, 2016.

\bibitem[\protect\citeauthoryear{Kim \bgroup \em et al.\egroup }{2021a}]{lab4}
Inhwan Kim, Sarvar~Hussain Nengroo, and Dongsoo Har.
\newblock Reinforcement learning for navigation of mobile robot with lidar.
\newblock In {\em 2021 5th International Conference on Electronics,
  Communication and Aerospace Technology (ICECA)}, pages 148--154. IEEE, 2021.

\bibitem[\protect\citeauthoryear{Kim \bgroup \em et al.\egroup }{2021b}]{lab7}
Taeyoung Kim, Luiz~Felipe Vecchietti, Kyujin Choi, Sanem Sariel, and Dongsoo
  Har.
\newblock Two-stage training algorithm for ai robot soccer.
\newblock {\em PeerJ Computer Science}, 7:e718, 2021.

\bibitem[\protect\citeauthoryear{Li \bgroup \em et al.\egroup
  }{2021}]{weight_factor5}
Mingchen Li, Xuechen Zhang, Christos Thrampoulidis, Jiasi Chen, and Samet
  Oymak.
\newblock Autobalance: Optimized loss functions for imbalanced data.
\newblock {\em Advances in Neural Information Processing Systems},
  34:3163--3177, 2021.

\bibitem[\protect\citeauthoryear{Li}{2017}]{deep_rl}
Yuxi Li.
\newblock Deep reinforcement learning: An overview.
\newblock {\em arXiv preprint arXiv:1701.07274}, 2017.

\bibitem[\protect\citeauthoryear{Lillicrap \bgroup \em et al.\egroup
  }{2015}]{ddpg}
Timothy~P Lillicrap, Jonathan~J Hunt, Alexander Pritzel, Nicolas Heess, Tom
  Erez, Yuval Tassa, David Silver, and Daan Wierstra.
\newblock Continuous control with deep reinforcement learning.
\newblock {\em arXiv preprint arXiv:1509.02971}, 2015.

\bibitem[\protect\citeauthoryear{Lin \bgroup \em et al.\egroup
  }{2020}]{skewed3}
Enlu Lin, Qiong Chen, and Xiaoming Qi.
\newblock Deep reinforcement learning for imbalanced classification.
\newblock {\em Applied Intelligence}, 50(8):2488--2502, 2020.

\bibitem[\protect\citeauthoryear{Lin}{1992}]{ER}
Long-Ji Lin.
\newblock Self-improving reactive agents based on reinforcement learning,
  planning and teaching.
\newblock {\em Machine learning}, 8(3):293--321, 1992.

\bibitem[\protect\citeauthoryear{Menon \bgroup \em et al.\egroup
  }{2013}]{weight_factor2}
Aditya Menon, Harikrishna Narasimhan, Shivani Agarwal, and Sanjay Chawla.
\newblock On the statistical consistency of algorithms for binary
  classification under class imbalance.
\newblock In {\em International Conference on Machine Learning}, pages
  603--611. PMLR, 2013.

\bibitem[\protect\citeauthoryear{Mnih \bgroup \em et al.\egroup }{2013}]{atrai}
Volodymyr Mnih, Koray Kavukcuoglu, David Silver, Alex Graves, Ioannis
  Antonoglou, Daan Wierstra, and Martin Riedmiller.
\newblock Playing atari with deep reinforcement learning.
\newblock {\em arXiv preprint arXiv:1312.5602}, 2013.

\bibitem[\protect\citeauthoryear{Moon \bgroup \em et al.\egroup }{2022}]{lab5}
Woohyeon Moon, Bumgeun Park, Sarvar~Hussain Nengroo, Taeyoung Kim, and Dongsoo
  Har.
\newblock Path planning of cleaning robot with reinforcement learning.
\newblock {\em arXiv preprint arXiv:2208.08211}, 2022.

\bibitem[\protect\citeauthoryear{Moore and
  Atkeson}{1993}]{prioritized_sweeping}
Andrew~W Moore and Christopher~G Atkeson.
\newblock Prioritized sweeping: Reinforcement learning with less data and less
  time.
\newblock {\em Machine learning}, 13(1):103--130, 1993.

\bibitem[\protect\citeauthoryear{Ng \bgroup \em et al.\egroup
  }{1999}]{reward_shaping}
Andrew~Y Ng, Daishi Harada, and Stuart Russell.
\newblock Policy invariance under reward transformations: Theory and
  application to reward shaping.
\newblock In {\em Icml}, volume~99, pages 278--287, 1999.

\bibitem[\protect\citeauthoryear{Novati and
  Koumoutsakos}{2019}]{remember_forget}
Guido Novati and Petros Koumoutsakos.
\newblock Remember and forget for experience replay.
\newblock In {\em International Conference on Machine Learning}, pages
  4851--4860. PMLR, 2019.

\bibitem[\protect\citeauthoryear{Park \bgroup \em et al.\egroup }{2022}]{lab8}
Bumgeun Park, Jihui Lee, Taeyoung Kim, and Dongsoo Har.
\newblock Kick-motion training with dqn in ai soccer environment.
\newblock {\em arXiv preprint arXiv:2212.00389}, 2022.

\bibitem[\protect\citeauthoryear{Schaul \bgroup \em et al.\egroup }{2015}]{PER}
Tom Schaul, John Quan, Ioannis Antonoglou, and David Silver.
\newblock Prioritized experience replay.
\newblock {\em arXiv preprint arXiv:1511.05952}, 2015.

\bibitem[\protect\citeauthoryear{Seo \bgroup \em et al.\egroup }{2019}]{lab1}
Minah Seo, Luiz~Felipe Vecchietti, Sangkeum Lee, and Dongsoo Har.
\newblock Rewards prediction-based credit assignment for reinforcement learning
  with sparse binary rewards.
\newblock {\em IEEE Access}, 7:118776--118791, 2019.

\bibitem[\protect\citeauthoryear{Shapiro \bgroup \em et al.\egroup
  }{2001}]{gaussian}
Linda~G Shapiro, George~C Stockman, et~al.
\newblock {\em Computer vision}, volume~3.
\newblock Prentice Hall New Jersey, 2001.

\bibitem[\protect\citeauthoryear{Silver \bgroup \em et al.\egroup }{2016}]{go1}
David Silver, Aja Huang, Chris~J Maddison, Arthur Guez, Laurent Sifre, George
  Van Den~Driessche, Julian Schrittwieser, Ioannis Antonoglou, Veda
  Panneershelvam, Marc Lanctot, et~al.
\newblock Mastering the game of go with deep neural networks and tree search.
\newblock {\em nature}, 529(7587):484--489, 2016.

\bibitem[\protect\citeauthoryear{Silver \bgroup \em et al.\egroup }{2017}]{go2}
David Silver, Julian Schrittwieser, Karen Simonyan, Ioannis Antonoglou, Aja
  Huang, Arthur Guez, Thomas Hubert, Lucas Baker, Matthew Lai, Adrian Bolton,
  et~al.
\newblock Mastering the game of go without human knowledge.
\newblock {\em nature}, 550(7676):354--359, 2017.

\bibitem[\protect\citeauthoryear{Sinha \bgroup \em et al.\egroup
  }{2022}]{weighting_to_loss}
Samarth Sinha, Jiaming Song, Animesh Garg, and Stefano Ermon.
\newblock Experience replay with likelihood-free importance weights.
\newblock In {\em Learning for Dynamics and Control Conference}, pages
  110--123. PMLR, 2022.

\bibitem[\protect\citeauthoryear{Sutton and Barto}{2018}]{RL}
Richard~S Sutton and Andrew~G Barto.
\newblock {\em Reinforcement learning: An introduction}.
\newblock MIT press, 2018.

\bibitem[\protect\citeauthoryear{Todorov \bgroup \em et al.\egroup
  }{2012}]{mujoco}
Emanuel Todorov, Tom Erez, and Yuval Tassa.
\newblock Mujoco: A physics engine for model-based control.
\newblock In {\em 2012 IEEE/RSJ international conference on intelligent robots
  and systems}, pages 5026--5033. IEEE, 2012.

\bibitem[\protect\citeauthoryear{Vecchietti \bgroup \em et al.\egroup
  }{2020a}]{lab2}
Luiz~Felipe Vecchietti, Taeyoung Kim, Kyujin Choi, Junhee Hong, and Dongsoo
  Har.
\newblock Batch prioritization in multigoal reinforcement learning.
\newblock {\em IEEE Access}, 8:137449--137461, 2020.

\bibitem[\protect\citeauthoryear{Vecchietti \bgroup \em et al.\egroup
  }{2020b}]{lab3}
Luiz~Felipe Vecchietti, Minah Seo, and Dongsoo Har.
\newblock Sampling rate decay in hindsight experience replay for robot control.
\newblock {\em IEEE Transactions on Cybernetics}, 2020.

\bibitem[\protect\citeauthoryear{White \bgroup \em et al.\egroup
  }{2014}]{surprise_data}
Adam White, Joseph Modayil, and Richard~S Sutton.
\newblock Surprise and curiosity for big data robotics.
\newblock In {\em Workshops at the Twenty-Eighth AAAI Conference on Artificial
  Intelligence}, 2014.

\bibitem[\protect\citeauthoryear{Zhang and Sutton}{2017}]{CER}
Shangtong Zhang and Richard~S Sutton.
\newblock A deeper look at experience replay.
\newblock {\em arXiv preprint arXiv:1712.01275}, 2017.

\bibitem[\protect\citeauthoryear{Zhang \bgroup \em et al.\egroup
  }{2017}]{skewed2}
Liangpeng Zhang, Ke~Tang, and Xin Yao.
\newblock Log-normality and skewness of estimated state/action values in
  reinforcement learning.
\newblock {\em Advances in Neural Information Processing Systems}, 30, 2017.

\bibitem[\protect\citeauthoryear{Zhang \bgroup \em et al.\egroup
  }{2020}]{weight_factor4}
Kun Zhang, Zhiyong Wu, Daode Yuan, Jian Luan, Jia Jia, Helen Meng, and Binheng
  Song.
\newblock Re-weighted interval loss for handling data imbalance problem of
  end-to-end keyword spotting.
\newblock In {\em INTERSPEECH}, volume 108, pages 2567--2571, 2020.

\end{thebibliography}

\end{document}